# Multi-focus Image Fusion Based on Similarity Characteristics

Ya-Qiong Zhang, Xiao-Jun Wu*, Hui Li
(School of IoT Engineering, Jiangnan University, Wuxi, 214122, China)
z_yaqiong@163.com, xiaojun_wu_jnu@163.com, hui_li_jnu@163.com

## Abstract

*A novel multi-focus image fusion algorithm performed in spatial domain based on similarity characteristics is proposed incorporating with region segmentation. In this paper, a new similarity measure is developed based on the structural similarity (SSIM) index, which is more suitable for multi-focus image segmentation. Firstly, the SSNSIM map is calculated between two input images. Then we segment the SSNSIM map using watershed method, and merge the small homogeneous regions with fuzzy c-means clustering algorithm (FCM). For three source images, a joint region segmentation method based on segmentation of two images is used to obtain the final segmentation result. Finally, the corresponding segmented regions of the source images are fused according to their average gradient. The performance of the image fusion method is evaluated by several criteria including spatial frequency, average gradient, entropy, edge retention etc. The evaluation results indicate that the proposed method is effective and has good visual perception.*

## 1. Introduction

Image fusion provides a mathematical model to integrate complementary and redundant information from multiple images of the same scene into a composite image, which is more suitable for the purpose of human visual perception and computer-processing tasks, such as segmentation, feature extraction and target recognition. Important applications of image fusion include medical science [1,27], remote sensing [2,17,18], face recognition [3] and object tracking [19,20].

In this paper we concentrate on multi-focus image fusion [4,5,16]. Due to the limited depth-of-focus of optical lenses, it is impossible to acquire an image that contains all relevant objects in focus through a shooting. To achieve an image where all objects are in focus, a multi-focus image fusion technique is desirable.

Image fusion is generally performed at three different levels of information representation including pixel level, feature level and decision level respectively. According to recent research [6-8,21,24,27], feature level image fusion based on regions has become an important method [22,23,25,26,29]. Several researchers incorporated the regional feature information in multi-resolution decomposition [9,10], while the inverse transform process may lose some information [11].

A recent trend for region based image fusion method is to implement the segmentation and fusion in spatial domain [11,12]. Li and Yang [11] segment the averaged image via traditional image segmentation without considering the complementary and redundant information, which is likely to produce segmentation mistakes. A partition strategy according to the *SSIM*-map of the input images has been proposed in [12], however, the segmentation result of multi-focus images is less reliable since different focus areas can't be clearly distinguished by SSIM.

Fig.1 depicts our proposed image fusion process based on similarity characteristics and region segmentation for two source images. It gives a description of our segmentation process: non-similarity based segmentation incorporating watershed and fuzzy c-means clustering algorithm FCM(NSIWF). Firstly, the SSIM map and the sign for different regions are calculated between two input images, and our signed structural non-similarity (SSNSIM) map is generated by combining SSIM and the sign. To obtain an accurate and reliable region segmentation result, we merge the small regions which are generated by watershed using FCM. After achieving the final segmentation result, the corresponding segmented regions of the source images are fused according to the fusion rule.

Fig.2 shows the fusion process of three source images based on joint region segmentation. The joint region segmentation is based on segmentation of two source images. Three different segmentation results are obtained by using NSIWF. To achieve a more reasonable final segmentation result that considering the information of all the source images, the joint region segmentation result is generated by integrating all of the three segmentation results.

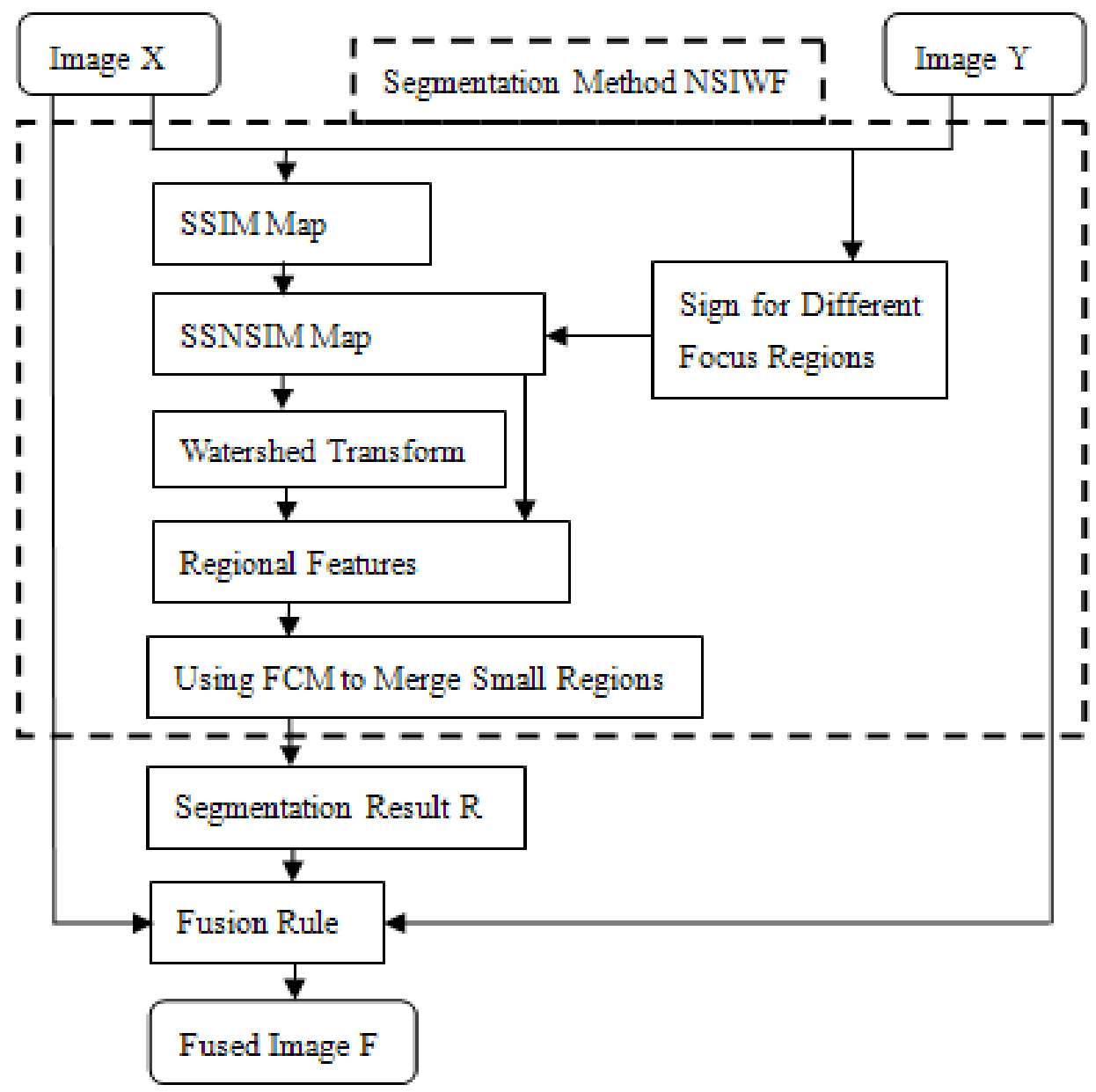

Figure 1: Fusion process for two source images

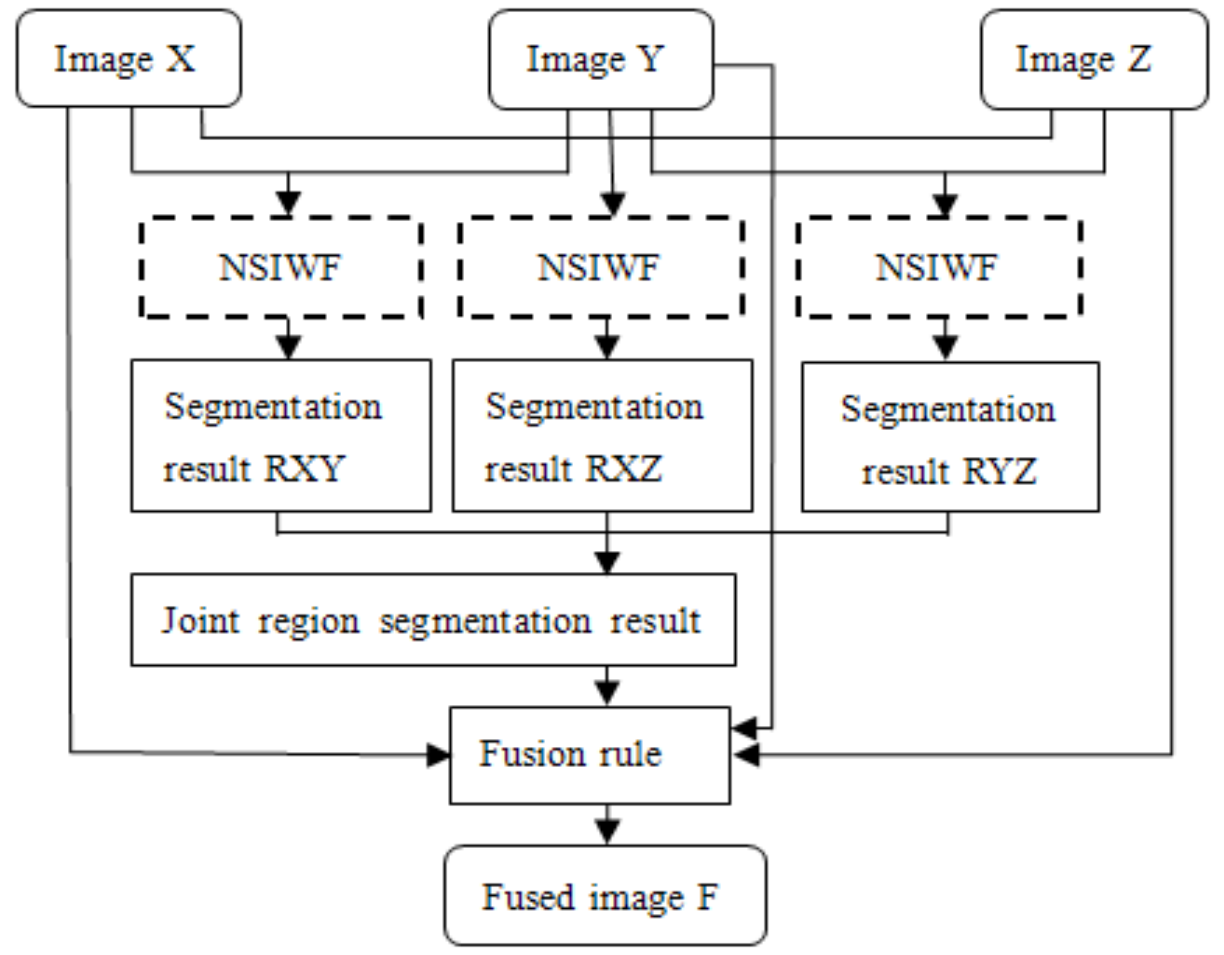

Figure 2: Fusion process for three source images

The main contributions of this paper include:

(1) A new similarity measure is developed: signed structural non-similarity (SSNSIM) index, which leads to a more reliable region segmentation process for multi-focus image fusion.

(2) A joint region segmentation method for three source images based on segmentation of two images is used to obtain a more accurate segmentation result.

The remainder of the paper is organized as follows. In Section 2, the proposed similarity measure SSNSIM is described. Section 3 provides a brief introduction to the segmentation method and joint region method. Our fusion rule is presented in Section 4. The experimental results are presented in Section 5 and conclusions are drawn in Section 6.

## 2. Similarity measures

### 2.1. Structural similarity index(SSIM)

It is necessary to analyze the similarity of source images, because the purpose of image fusion is to create a composite image, which preserves the complementary characteristics and removes the redundant information. *SSIM* has been used as similarity index in [12].

Given two images $X$ and $Y$, the *SSIM* value for each pixel is evaluated as [13]:

$$SSIM(i,j) = [l(i,j)]^{\alpha}[c(i,j)]^{\beta}[s(i,j)]^{\gamma} \quad (1)$$

where $\alpha > 0$, $\beta > 0$ and $\gamma > 0$ are parameters used to adjust the relative importance of the three components. The luminance $l(i,j)$, contrast $c(i,j)$, and structure $s(i,j)$ components for each pixel are defined as follows:

$$l(i,j) = \frac{2\mu_X(i,j)\mu_Y(i,j) + C_1}{\mu_X(i,j)^2 + \mu_Y(i,j)^2 + C_1} \quad (2)$$

$$c(i,j) = \frac{2\sigma_X(i,j)\sigma_Y(i,j) + C_2}{\sigma_X(i,j)^2 + \sigma_Y(i,j)^2 + C_2} \quad (3)$$

$$s(i,j) = \frac{\sigma_{XY}(i,j) + C_3}{\sigma_X(i,j)\sigma_Y(i,j) + C_3} \quad (4)$$

where $\mu_X(i,j)$ and $\mu_Y(i,j)$ are the means of central pixel and neighborhood pixels in the local window for $X$ and $Y$ respectively. $\sigma_X(i,j)$ and $\sigma_Y(i,j)$ are the standard deviation between central pixel and neighborhood pixels in the local window for $X$ and $Y$ respectively. $\sigma_{XY}(i,j)$ is the cross-correlation of $X - \mu_X(i,j)$ and $Y - \mu_y(i,j)$ in the local window centered on pixel $(i,j)$ for $X$ and $Y$.

### 2.2. Signed structural non-similarity(SSNSIM)

The *SSIM* has already been applied to region segmentation [12]. As shown in Fig.3(c), different focus areas are not be clearly distinguished by *SSIM*-map, which would reduce the accuracy of segmentation and fusion performance.

Generally, the similarity between two source images in different focus areas is small, while the non-similarity is large. To make the feature difference between the two kinds of regions more apparent, a new similarity measure is developed based on *SSIM* by appending sign for different focus areas: signed structural non-similarity (*SSNSIM*) index.

Let $SNSIM$ be the structural non-similarity, the $SNSIM$ value for each pixel is:

$$SNSIM(i,j) = 1 - SSIM(i,j) \quad (6)$$

Let $SSNSIM$ be the signed structural non-similarity, the $SSNSIM$ value for each pixel is defined as:

$$SSNSIM(i,j) = S(i,j)SNSIM(i,j) \quad (7)$$

where $S$ is the sign for different focus areas, defined by $\sigma_X(i,j)$ and $\sigma_Y(i,j)$ in (3):

$$S(i,j) = \begin{cases} +1, \sigma_X(i,j) > \sigma_Y(i,j) \\ -1, \sigma_X(i,j) \le \sigma_Y(i,j) \end{cases} \quad (8)$$

where +1 and −1 denote the pixel belonging to the focus areas in image $X$ and image $Y$, respectively. In apparently different focus areas, the $SNSIM$ is large and the sign $S$ is opposite, thus $SSNSIM$ are very different. Fig.3(d) shows the *SSNSIM*-map for two source images, higher gray value indicates larger $SSNSIM$ value. As it can be seen, different focus areas are clearly distinguished. Thus, the use of SSNSIM can lead to a more reliable segmentation process.

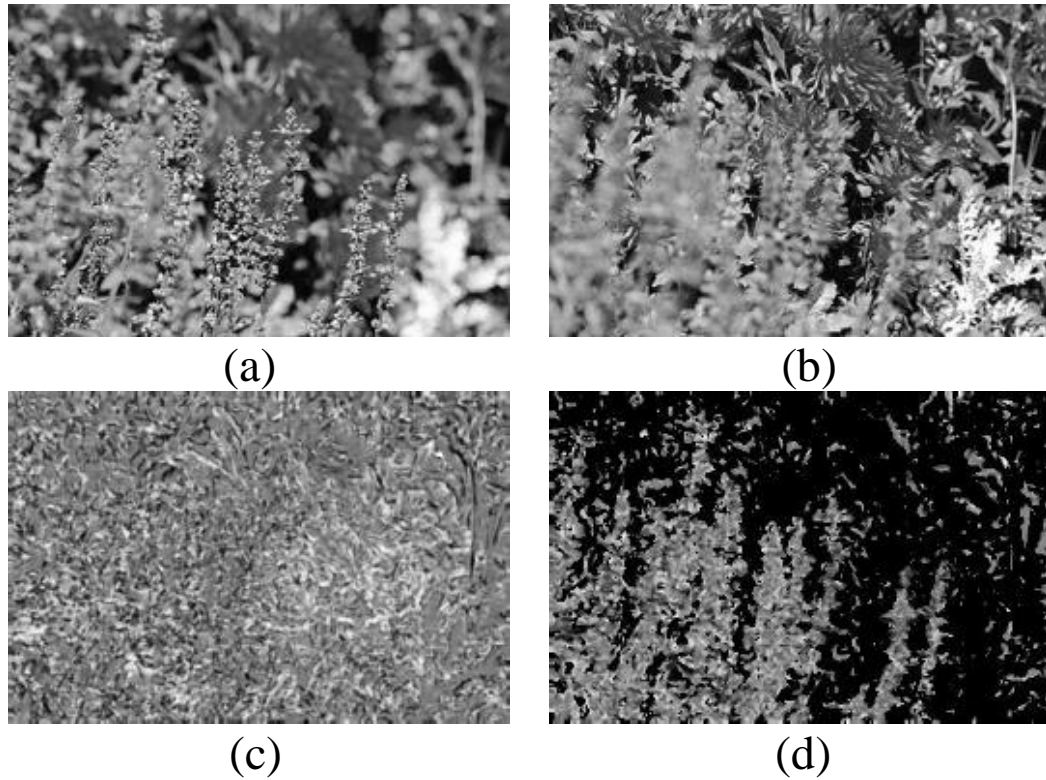

Figure 3: Similarity maps of multi-focus images (a) and (b). (c) *SSIM*-map. (d) *SSNSIM*-map.

## 3. Image segmentation

### 3.1. Segmentation based on watershed and FCM

Different source image regions have different similarity so that the similarity correlation is regional. Therefore, the similarity map should be partitioned into regions.

The watershed algorithm is a common and powerful technique for image segmentation, and has recently received increased attention for its high segmentation accuracy and low computational load. However, this transformation is sensitive to noise, and usually results in over segmentation. The over segmentation makes it hard to identify meaningful regions.

To reduce the over segmentation, firstly, the morphological H-minima transform [14] is performed on the gradient map of *SSNSIM*-map as a modification, then the watershed algorithm is performed on the modified gradient map. Two sub-images at the top-left $50 \times 50$ pixels of Fig.3 are exemplified in Fig.4 to illustrate the process of forming regions better. The segmentation output of the watershed algorithm is shown in Fig.4(a).

To deal with the over segmented output of the watershed algorithm, we need to merge the over segmented regions. Fuzzy c-means clustering algorithm (FCM) is an effective solution to this problem.

Let $num$ be the number of regions segmented by watershed algorithm and the SSNSIM values for the $n$-th region is $SSNSIM_n (1 \le n \le num)$, we use the mean SSNSIM (MNS) as the data for clustering:

$$MNS_n = \frac{1}{M_n} \sum_{k=1}^{M_n} SSNSIM_n(k) \quad (9)$$

where $M_n$ is the number of pixels in the $n$-th region.

Let $N_{cluster}$ be the number of clusters. The FCM algorithm is used to cluster the feature $MNS_n (1 \le n \le num)$, we obtain $\mu_{mn} (1 \le m \le N_{cluster}, 1 \le n \le num)$, $\mu_{mn} \in [0,1]$ denotes the membership of the feature $MNS_n$ in the $m$-th cluster. The classification strategy is given as:

$MNS_n$ belongs to the $m$-th cluster when $\mu_{mn} = \max_l \{\mu_{ln}, l = 1,2,\ldots,N_{cluster}\}$. The feature $MNS_n$ is partitioned into $N_{cluster}$ groups, and consequently, regions segmented by watershed algorithm are partitioned into $N_{cluster}$ classes. Fig.4(b) shows the clustering result of FCM when $N_{cluster}$ is 5, where the regions of the same class are labeled by the same grey value.

Then, we merge the regions by dealing with the regional boundary as follows: the boundaries of the regions belong to the same class can be directly included in this class, and the boundaries of the different classes will be included in the class to which its feature is most close. Fig.4(c) shows the final segmentation result.

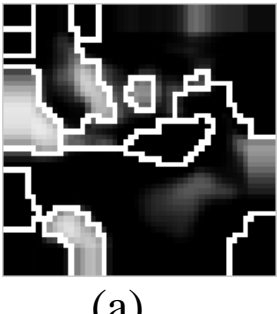
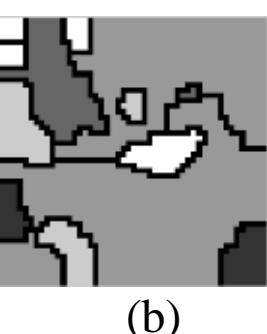
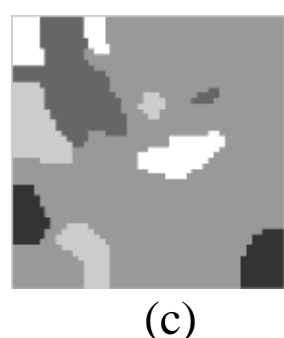

Figure 4: An example of region segmentation. (a) Segmentation output of the watershed algorithm. (b) Clustering result of FCM. (c) Final segmentation result.

### 3.2. Joint region segmentation

For the fusion of three source images, we proposed a joint region segmentation method based on the similarity between all the source images.

Given three images $X$, $Y$ and $Z$, $R_{XY}$ is the segmentation result obtained with $X$ and $Y$, $R_{XZ}$ is the segmentation result obtained with $X$ and $Z$, $R_{YZ}$ is the segmentation result obtained with $Y$ and $Z$. Joint region segmentation is a process to incorporate $R_{XY}$, $R_{XZ}$ and $R_{YZ}$, and generates a more accurate segmentation result. Fig.5 shows an example of generating a joint region map of $R_{XY}$ and $R_{XZ}$. Let $r_{XY}$ denotes a region from $R_{XY}$, and $r_{XZ}$ denotes a region from $R_{XZ}$, we set the following incorporation rules:

(1) If $r_{XY}$ is the same as $r_{XZ}$, one region is generated in the joint region map: $r_J^1 = r_{XY} = r_{XZ}$.

(2) If $r_{XY}$ is disjoint with $r_{XZ}$, two regions are generated in the joint region map: $r_J^1 = r_{XY}$, $r_J^2 = r_{XZ}$.

(3) If one region is entirely included in the other one, e.g. $r_{XY} \subset r_{XZ}$, two regions are generated in the joint region map: $r_J^1 = r_{XY}$, $r_J^2 = r_{XZ} - r_{XY}$.

(4) If $r_{XY}$ and $r_{XZ}$ are partially overlapped, three regions are produced in the joint region map: $r_J^1 = r_{XY} \cap r_{XZ}$, $r_J^2 = r_{XY} - r_J^1$, $r_J^3 = r_{XZ} - r_J^1$.

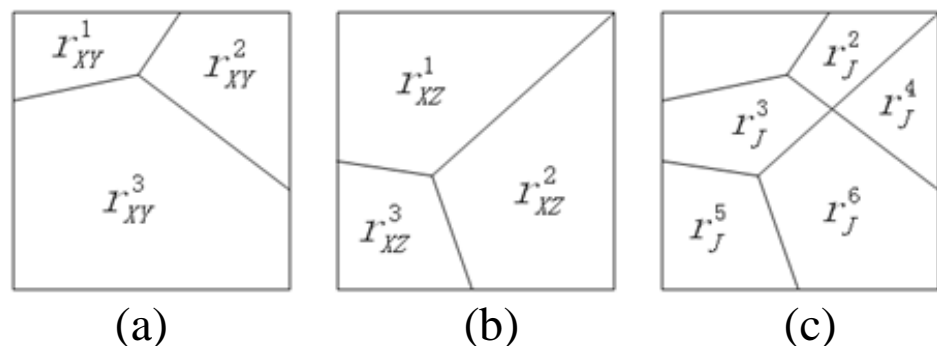

Figure 5: An example of generating a joint region map of $R_{XY}$ and $R_{XZ}$. (a) $R_{XY}$. (b) $R_{XZ}$. (c) Joint region map $R_J$.

The final segmentation result $R_{XYZ}$ is obtained by generating a joint region map of $R_J$ and $R_{XZ}$.

In this paper, we use a simple way to implement the joint region segmentation according to the above rules. Assuming $R_{XY}$, $R_{XZ}$ and $R_{YZ}$ are obtained by the FCM with a clustering number $N$, the joint region segmentation procedure is realized as follows:

(1) The regions of $R_{XY}$, $R_{XZ}$ and $R_{YZ}$ are labeled by the following three arithmetic progression respectively: $\{1, 2, \ldots, N\}$, $\{N+1, 2(N+1), \ldots, N(N+1)\}$, $\{(N+1)^2, 2(N+1)^2, \ldots, N(N+1)^2\}$. Assuming $A$, $B$ and $C$ are the label matrixes of $R_{XY}$, $R_{XZ}$ and $R_{YZ}$, which is shown in Fig.6.

(2) The label matrix of joint regions is:

$$M_{ABC} = A + B + C \tag{10}$$

Fig.6(d) is the label-map of joint regions, in which every number labels a region. The label matrix $M_{ABC}$ indicates the final segmentation result $R_{XYZ}$.

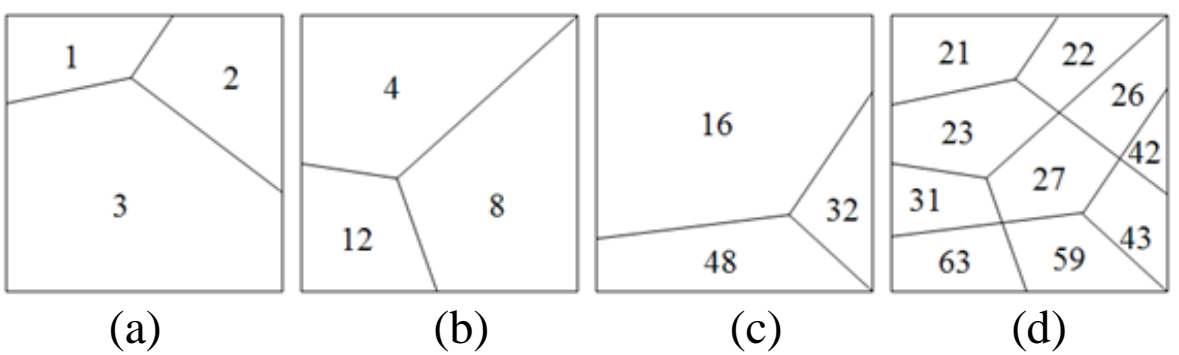

Figure 6: An example of joint region segmentation when the clustering number is 3. (a) Label matrix $A$ of $R_{XY}$. (b) Label matrix $B$ of $R_{XZ}$. (c) Label matrix $C$ of $R_{YZ}$. (d) Label matrix $M_{ABC}$ for $R_{XYZ}$.

## 4. Fusion rule

Here we adopt an average gradient sensitivity to evaluate the clarity of the image, it can reflect the features of image contrast in minute details and texture. The average gradient is defined as follows:

$$g = \frac{1}{(M-1)(N-1)} \sum_{i=1}^{M-1} \sum_{j=1}^{N-1} \sqrt{\frac{(F(i,j) - F(i+1,j))^2 + (F(i,j) - F(i,j+1))^2}{2}} \tag{11}$$

where $F$ is the fused image, $M \times N$ is the size of the fused image.

We compare the average gradient of the corresponding regions of the source images to decide which should be used to construct the fused image. Let's take the fusion of three source images as an example:

$$RoF_i = \begin{cases} RoX_i, g_i^X \geq g_i^Y and g_i^X \geq g_i^z \\ RoY_i, g_i^Y \geq g_i^X and g_i^Y \geq g_i^z \\ RoZ_i, g_i^Z \geq g_i^X and g_i^Z \geq g_i^Y \end{cases} \tag{12}$$

where $RoF_i$ is the $i$-th region of the fused image, $g_i^X$, $g_i^Y$ and $g_i^Z$ are the average gradients of the $i$-th region of image $X$, $Y$ and $Z$, respectively.

## 5. Experimental results and discussions

In this section, subjective tests and objective measurements are used for qualitatively and quantitatively assessing the performance of the proposed methodology in two source images fusion and three source images fusion respectively. The evaluation criteria include variance (V), spatial frequency (SF), average gradient (AG), entropy (H), mutual information (MI) and edge retention ($Q^{AB/F}$) [15].

These metrics can be effectively used to measure how well the source images are integrated into the fused image via the fusion algorithm. For all these measurements, the larger values indicate the better fusion performance.

The fused images of different fusion schemes on two source images are shown in Fig.7. Fig.7(c, d) show the segmentation results in different schemes, in which the same kind of regions are labeled by the same grey value. Fig.7(c) shows the segmentation result with RSIM [12], from which we can clearly see the defect that regions in different focus areas are under the same class. While our segmentation result in Fig.7(d) has better segmentation performance. Table 1 shows the corresponding objective evaluation, as it can be seen, our proposed method is inferior to RSIM only in MI compared with RSIM [12] and has better visual perception.

Table 1: Performance for Different Fusion Schemes with Fig.7

| Scheme | Evaluation | | | | |
|---|---|---|---|---|---|
| | V | SF | AG | H | MI |
| RSIM [12] | 51.3127 | 23.5837 | 12.0317 | 7.5888 | **1.6274** |
| Proposed | **52.8228** | **27.0478** | **13.8747** | **7.6143** | 1.5073 |

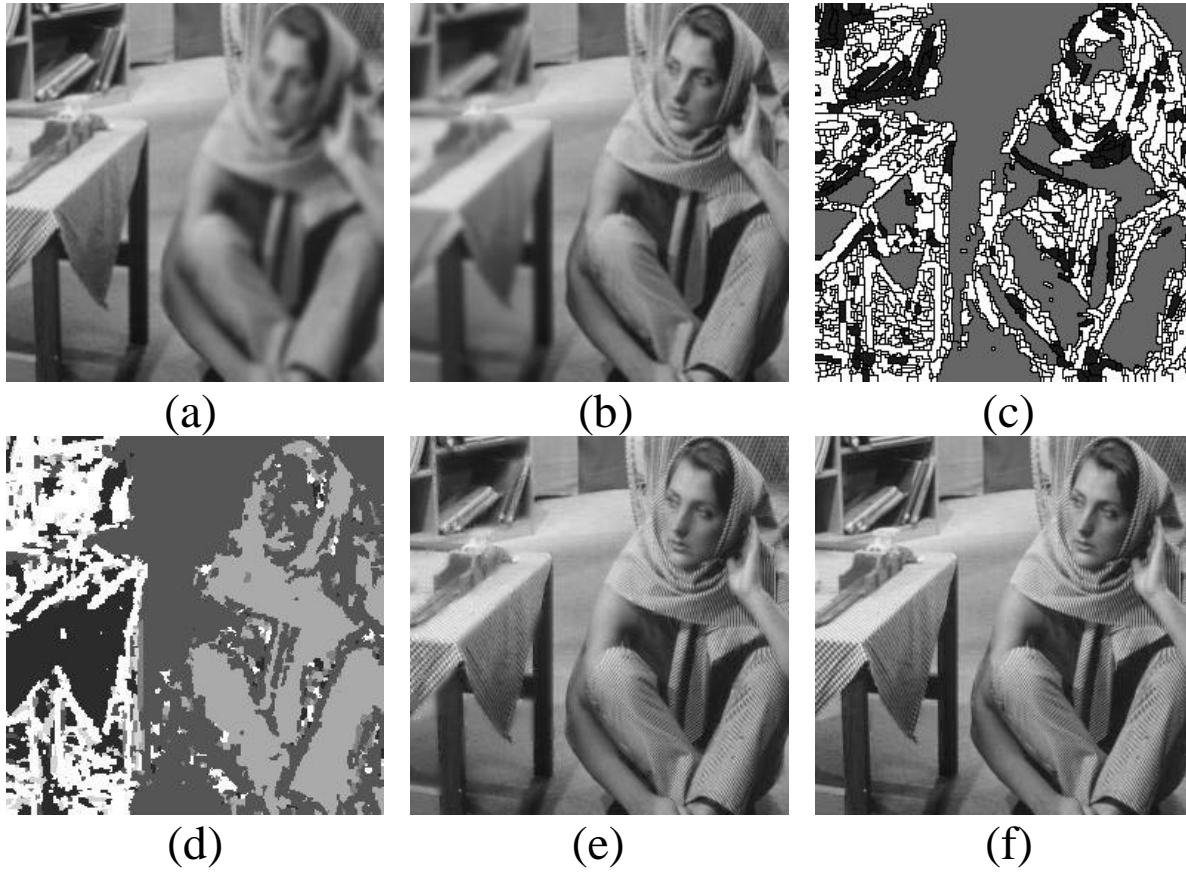

Figure 7: (a) Left focus image. (b) Right focus image. (c) Segmentation result with RSIM [12]. (d) Segmentation result with our proposed method. (e) Fused image with RSIM [12]. (f) Fused image with proposed method.

In order to test the robustness of our proposed algorithm, we simulate on other four pairs of two-focus images as shown in the columns of Fig.8(a) and (b). The fusion results of RSIM [12] and our proposed method are presented in the columns of Fig.8(c) and Fig.8(d), respectively.

Table 2 shows the objective comparison of different fusion schemes for the four pairs of original images in Fig.8. As we can see from Table 2, our method is inferior to RSIM [12] only in H for the first row of source images, which indicates that our method out performs RSIM [12] in two-focus image fusion.

Table 2: Performance for Different Fusion Schemes with Fig.8

| **Source images (row number in Fig.8)** | Scheme | Evaluation | | | | |
|---|---|---|---|---|---|---|
| | | V | SF | AG | H | MI |
| 1 | RSIM [12] | 51.0066 | 16.0091 | 6.6301 | **7.3274** | 1.4832 |
| | Proposed | **52.6560** | **18.4962** | **7.5417** | 7.3229 | **1.4865** |
| 2 | RSIM [12] | 47.8008 | 15.7682 | 8.6477 | 7.4323 | 1.3983 |
| | Proposed | **48.5790** | **18.6085** | **10.0506** | **7.4613** | **1.4108** |
| 3 | RSIM [12] | 53.1642 | 23.7357 | 11.0372 | 7.5090 | 1.5211 |
| | Proposed | **54.2567** | **26.7609** | **12.5070** | **7.5409** | **1.5332** |
| 4 | RSIM [12] | 46.5970 | 17.4105 | 6.7841 | 7.0250 | 1.6404 |
| | Proposed | **47.0912** | **19.0000** | **7.4256** | **7.0551** | **1.6444** |

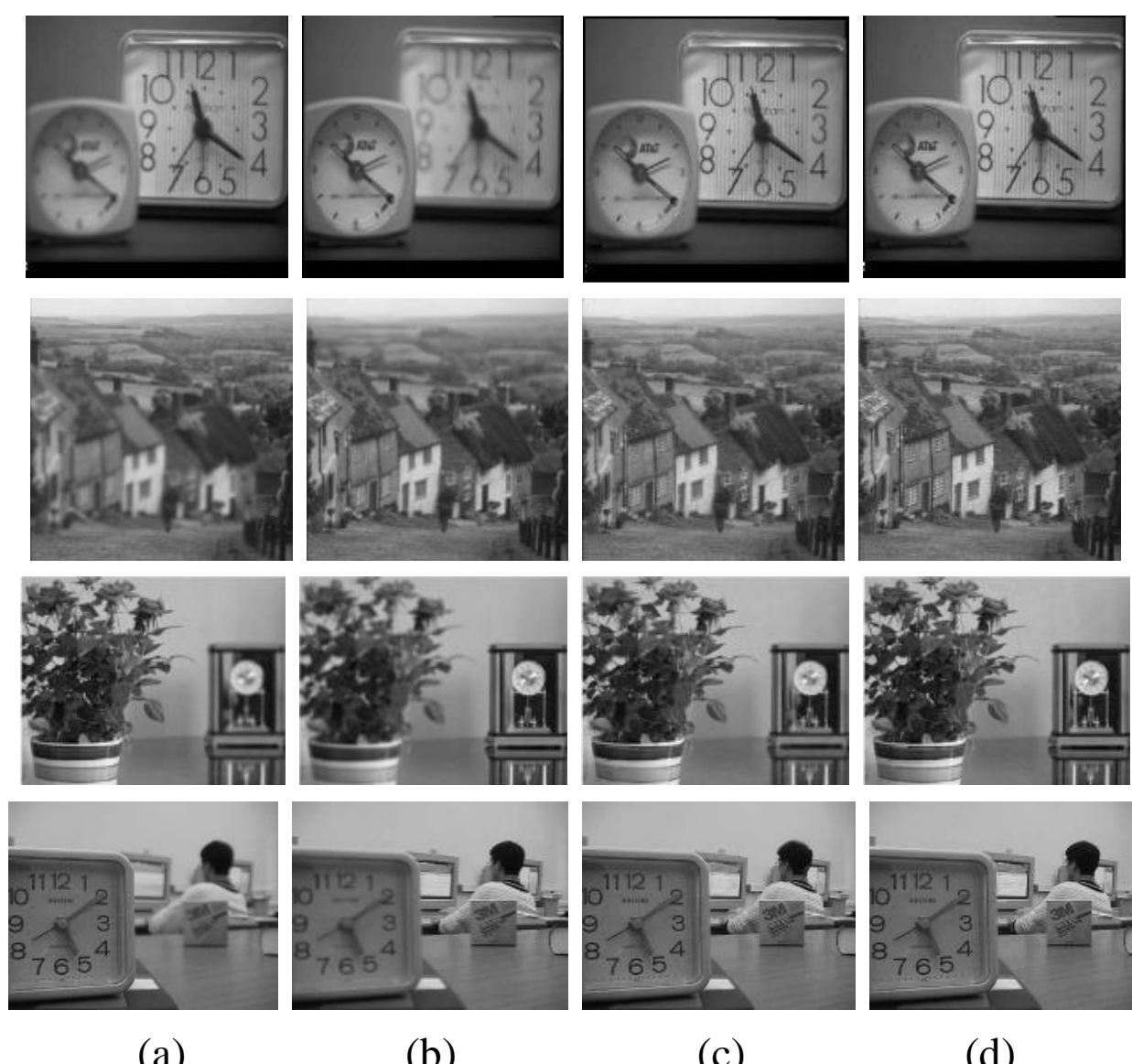

Figure 8: Fusion results of other images. (a) and (b) Source images. (c) Fused image by RSIM [12]. (d) Fused image by our proposed method.

Table 3 shows the effect of different fusion schemes on three source images in Fig.9 and Fig.10. The rectangular area in Fig.9 (e) indicates selecting fuzzy area of the source images. Fig.10 (j) shows the fused image with RSSF [11], in which the fuzzy part in the white circle indicates the segmentation error in the black circle of Fig.10(i). As it can be seen, our proposed method is more effective compared with RSSF [11] in both visual presentation and objective evaluation.

Table 3: Performance for different fusion schemes

| Image | Scheme | Evaluation | | | |
|---|---|---|---|---|---|
| | | SF | AG | H | $Q^{AB/F}$ |
| Fig.9 | RSSF [11] | 18.8296 | 7.7765 | 6.8012 | 0.6805 |
| | Proposed | **19.4607** | **8.3876** | **6.8161** | **0.6840** |
| Fig.10 | RSSF [11] | 24.4513 | 11.7633 | 7.3854 | 0.4929 |
| | Proposed | **25.0760** | **12.4874** | **7.4014** | **0.5134** |

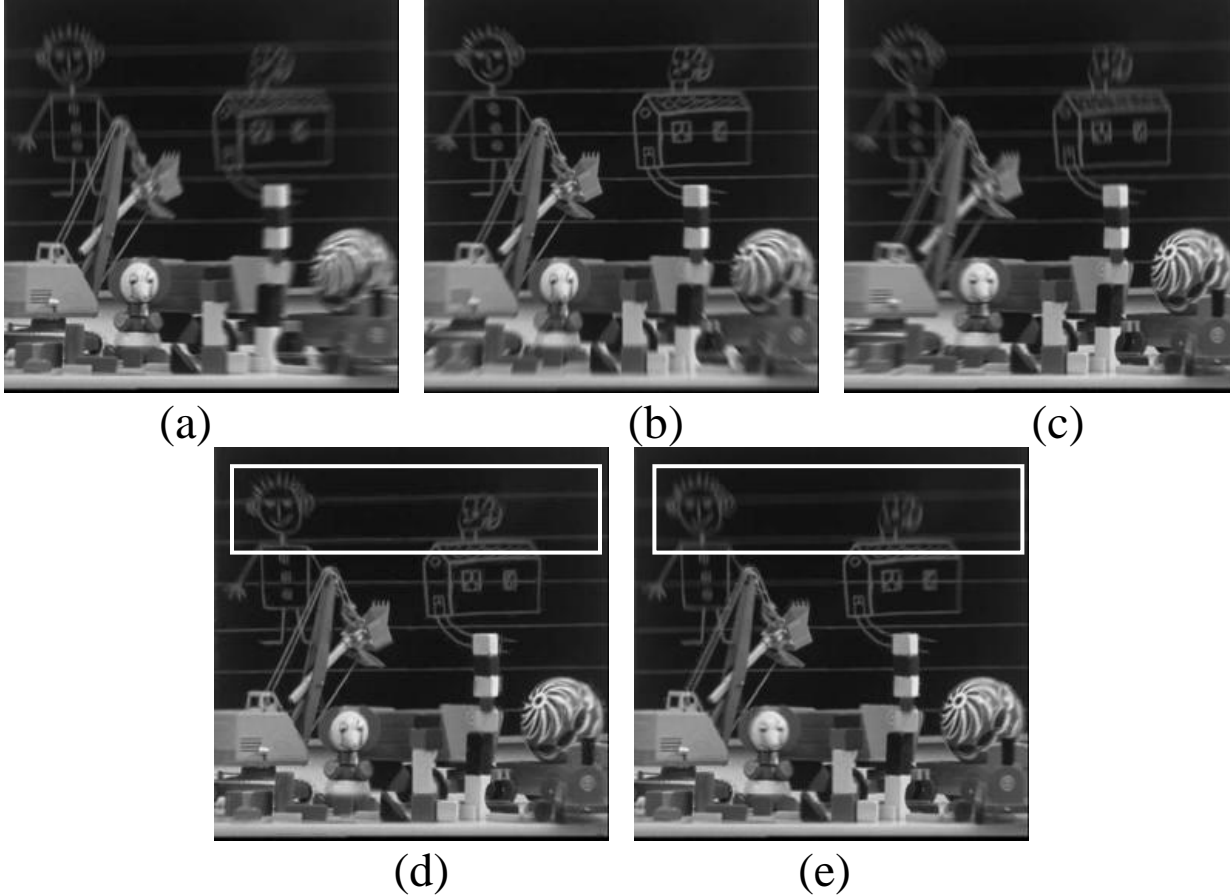

Figure 9: (a)-(c) Three source images. (d) Fused image with proposed method. (e) Fused image with RSSF [11].

Fig.10 (d-f) shows the fused image with segmentation result obtained by only two of the three source images, the fused image with joint segmentation result is shown in Fig.10 (h) and has better visual perception. Table 4 shows the effect of different segmentation results on three source images. As it can be seen, our joint region segmentation method is more reasonable and accurate and lead to a better fusion performance.

Table 4: Performance for different segmentation results

| Image | Segmentation result | Evaluation | | | |
|---|---|---|---|---|---|
| | | V | SF | AG | $Q^{AB/F}$ |
| Fig.10 | A | 23.9290 | 11.5754 | 7.3949 | 0.5071 |
| | B | 23.0867 | 11.4341 | **7.4016** | 0.4880 |
| | C | 24.3759 | 11.8802 | 7.3898 | 0.5084 |
| | $M_{ABC}$ | **25.0760** | **12.4874** | 7.4014 | **0.5134** |

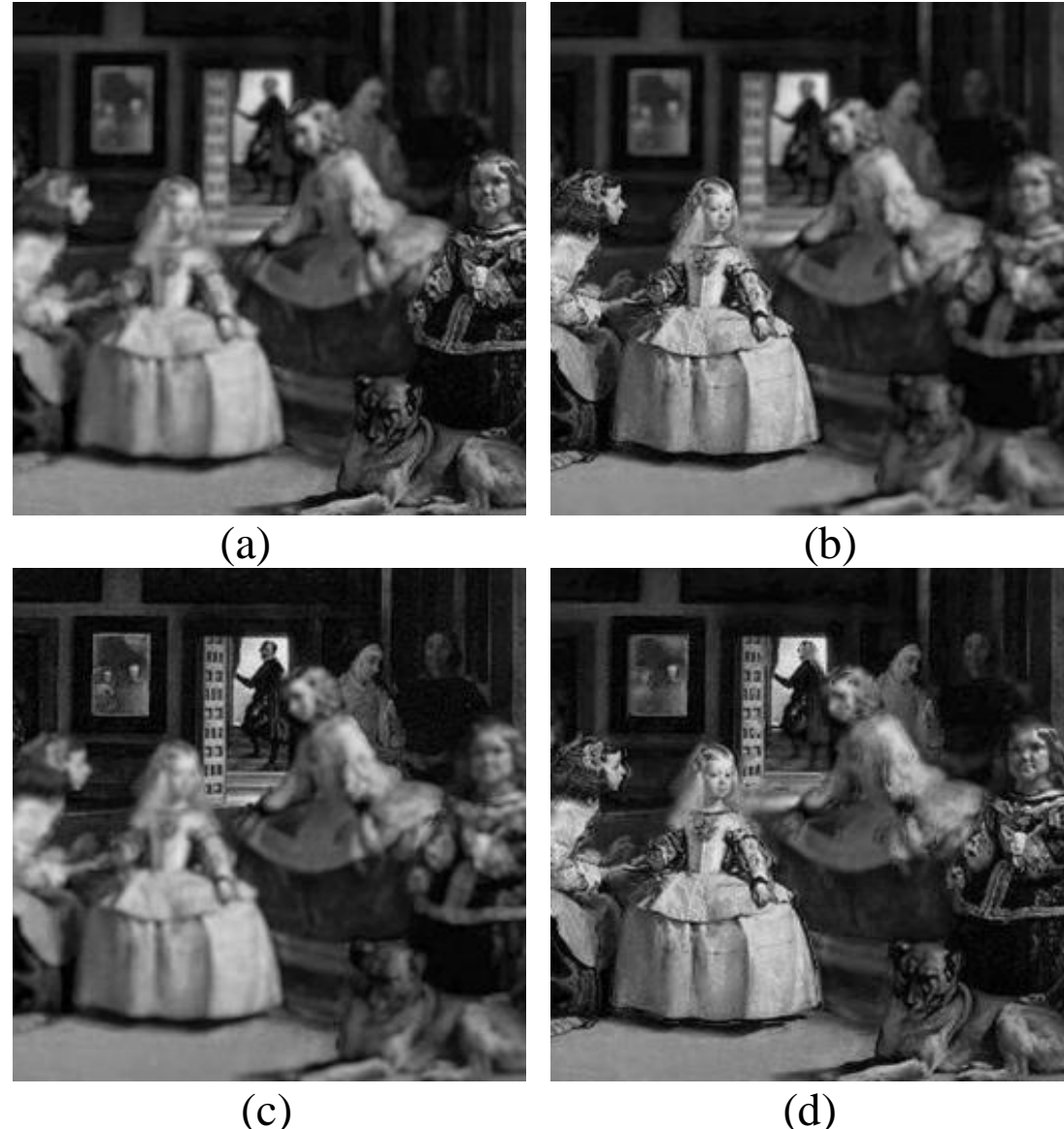

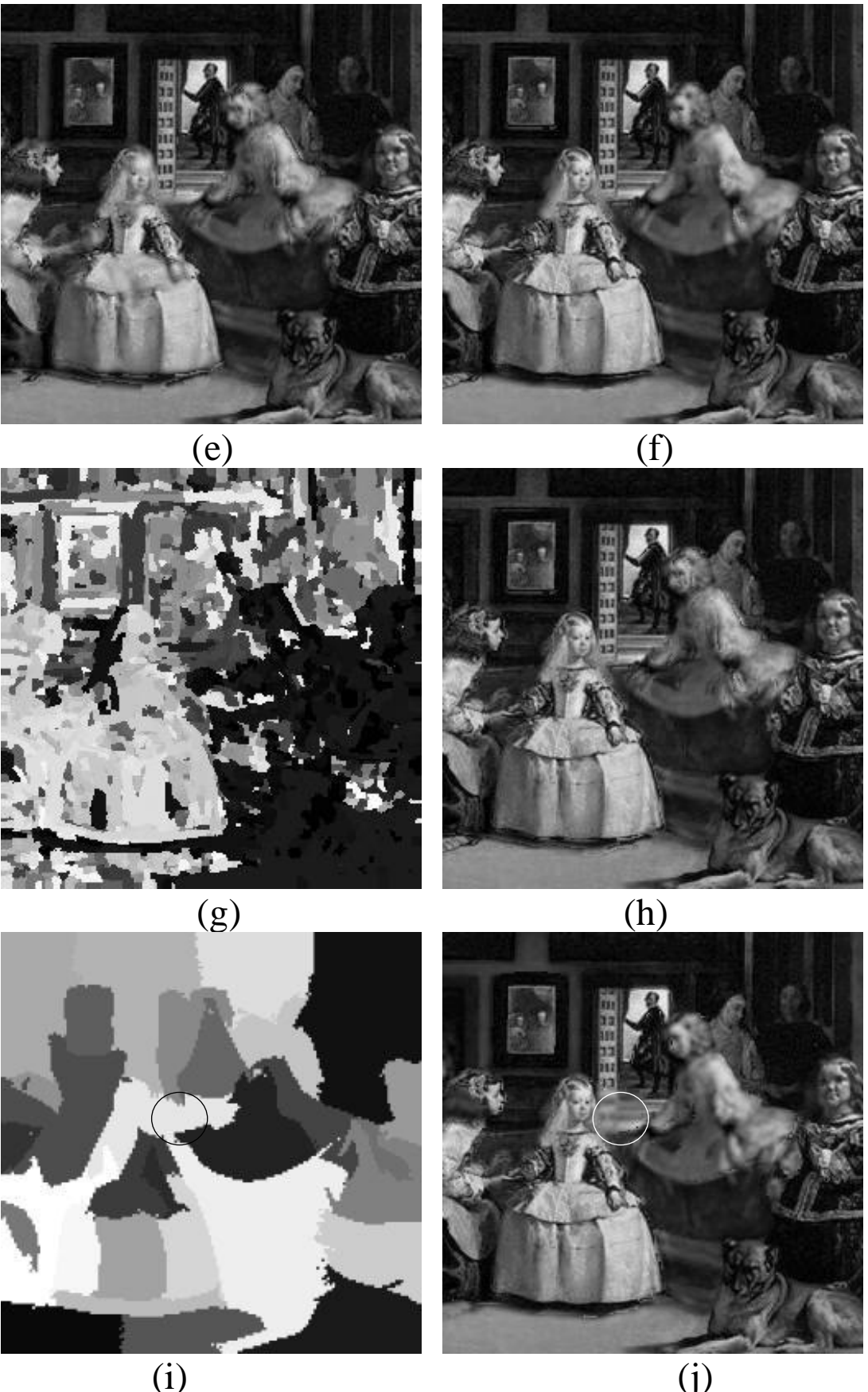

Figure 10: (a)-(c) Three source images. (d) Fused image with segmentation result obtained by (a) and (b). (e) Fused image with segmentation result obtained by (a) and (c). (f) Fused image with segmentation result obtained by (b) and (c). (g) Joint segmentation result. (h) Fused image with joint segmentation result. (i) Segmentation result with RSSF [11]. (j) Fused image with RSSF [11].

## 6. Conclusions

In this paper, we proposed a new multi-focus image fusion method performed in spatial domain based on similarity characteristics incorporating with region segmentation. Our proposed SSNSIM leads to a more reliable region segmentation process for multi-focus image fusion. Considering the complementary and redundant information between all the source images, our joint region segmentation method for three source images based on segmentation of two images generates a more accurate segmentation result. Experimental results conducted several groups of images show the proposed method is effective and the perceptive quality of the fused image is enhanced.